\documentclass{article}
\usepackage{spconf,amsmath,graphicx}
\usepackage{cleveref}
\usepackage{cite}
\usepackage{amsmath}
\usepackage{amssymb}
\usepackage{multirow}
\usepackage{enumitem}
\setlist{nosep, leftmargin=14pt}

\usepackage{mwe} % to get dummy images

\title{Medical Image Quality Assessment \\ based on Probability of Necessity and Sufficiency}
\name{Boyu Chen\textsuperscript{\rm 1}
, Ameenat L. Solebo\textsuperscript{\rm 2}, Weiye Bao\textsuperscript{\rm 3}, Paul Taylor\textsuperscript{\rm 1}
}

\address{
\textsuperscript{\rm 1}Institute of Health Informatics, University College London, UK\\
\textsuperscript{\rm 2}Great Ormond Street Institute of Child Health, University College London, UK\\
\textsuperscript{\rm 3}Division of Biosciences, University College London, UK\\
}
\begin{document}
%\ninept
%
\maketitle

\begin{abstract}
Medical image quality assessment (MIQA) is essential for reliable medical image analysis. While deep learning has shown promise in this field, current models could be misled by spurious correlations learned from data and struggle with out-of-distribution (OOD) scenarios. To that end, we propose an MIQA framework based on a concept from causal inference: Probability of Necessity and Sufficiency (PNS). PNS measures how likely a set of features is to be both necessary (always present for an outcome) and sufficient (capable of guaranteeing an outcome) for a particular result. Our approach leverages this concept by learning hidden features from medical images with high PNS values for quality prediction. This encourages models to capture more essential predictive information, enhancing their robustness to OOD scenarios. We evaluate our framework on an Anterior Segment Optical Coherence Tomography (AS-OCT) dataset for the MIQA task and experimental results demonstrate the effectiveness of our framework.

\end{abstract}

\section{Introduction}
Medical image quality assessment (MIQA) is crucial for reliable medical image analysis. While deep learning (DL) models have shown remarkable potential in automating MIQA \cite{chen2023uno, liu2023hierarchical, lin2023domain, petashvili2024learning, song2024md}, they face two challenges. First, DL models may capture features unrelated to quality from training data \cite{yang2023invariant}, potentially compromising MIQA performance. Second, DL models often struggle with out-of-distribution (OOD) scenarios \cite{fang2020rethinking}, a common challenge in medical imaging. They may fail to assess the quality of images that deviate from training data.
 % \cite{scholkopf2021toward}

To address these challenges, we turn to causal feature learning, a field that focuses on identifying features with causal influence on outcomes. Specifically, we use Probability of Necessity and Sufficiency (PNS) \cite{pearl2009causality}, detailed in \cref{sec:pns}, since recent work by Yang et al. \cite{yang2023invariant} has demonstrated PNS's potential to improve representation learning and mitigate OOD problems. In the context of MIQA, high-PNS features learned from images are more likely to have a causal link to quality, leading to accurate and robust predictions.

Inspired by this insight, we develop an MIQA framework to guide DL models to extract features with high PNS values for predicting good quality images. We evaluate our framework on an Anterior Segment Optical Coherence Tomography (AS-OCT) dataset for the MIQA task. Results demonstrate our framework can improve DL models in both predictive performance and OOD generalization in the MIQA task.

% The remainder of this paper is organized as follows: \cref{sec:pns} introduces the concept of PNS in detail, \cref{sec:method} describes how our framework extends PNS into feature extraction for MIQA, \cref{sec:exp} presents and discusses our experimental results, and \cref{sec:conclusion} concludes the paper.

\section{PNS}
\label{sec:pns}
\begin{figure}
\centerline{\includegraphics[width=\columnwidth]{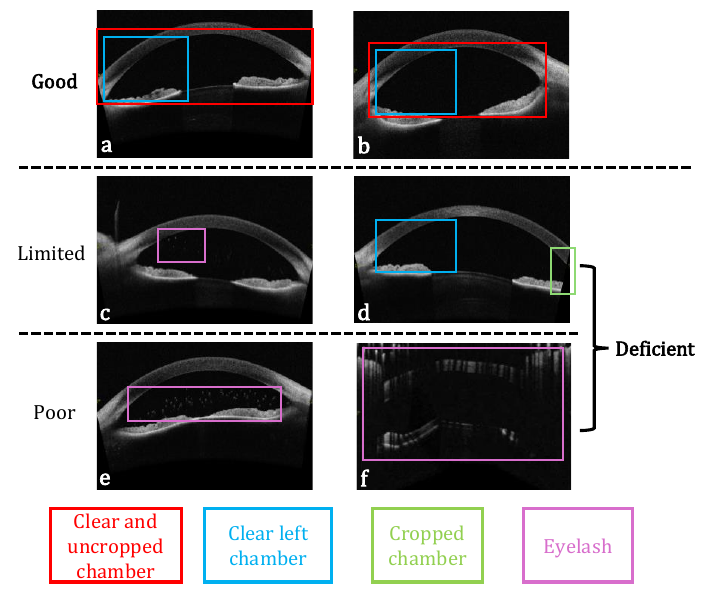}}
\caption{Quality rating of AS-OCT image.}
\label{fig:quality}
\end{figure}

PNS \cite{pearl2009causality} measures the likelihood of a feature set being both necessary and sufficient for an outcome. A feature is necessary if it is indispensable for an outcome, and sufficient if it alone can ensure the outcome. To illustrate this concept, consider MIQA for AS-OCT. As shown in \cref{fig:quality}, the quality is categorized as Good (complete, clear view of an eye chamber), Limited (relatively clear but with reduced fidelity, possibly having a cropped chamber or some artifacts like eyelashes), or Poor (unclear chamber blocked by artifacts like eyelashes) \cite{chen2023automated}. We group Limited and Poor into a single Deficient category to focus on distinguishing Good images. Then, ``Having a clear left chamber" is necessary for Good (\cref{fig:quality}. a/b) but not sufficient, as a cropped right chamber can still result in Deficient (\cref{fig:quality}. c). ``Having an eyelash artifact" is sufficient for Deficient (\cref{fig:quality}. c/e/f) but not necessary, as a cropped chamber without an eyelash artifact can also lead to Deficient (\cref{fig:quality}. d). Ideally, MIQA models should capture features that are both necessary and sufficient for Good, such as ``a clear and uncropped chamber" (\cref{fig:quality}. a/b).

Let $H$ be the hidden features of outcome $Y$. Let $h$ be an implementation of $H$ and its complement be $\bar{h}$. The PNS of $H$ with respect to $Y$ on $h$ and $\bar{h}$ is:
\begin{equation} \nonumber
\begin{aligned}
&\text{PNS}(Y, H) := \\
&P(Y_{\mathrm{do}(H=h)}=y|H=\bar{h},Y \neq y)P(H=\bar{h},Y \neq y) \\
&+ P(Y_{\mathrm{do}(H=\bar{h})}\neq y|H=h,Y=y)P(H=h,Y=y)\\
\end{aligned}
\end{equation}

In this equation, $P(Y_{\mathrm{do}(H=h)}=y|H=\bar{h},Y \neq y)$ represents the counterfactual probability of sufficiency. It is the probability of $Y=y$ if we were to intervene and set $H=h$, given that we actually observed $H=\bar{h}$ and $Y \neq y$. Similarly, $P(Y_{\mathrm{do}(H=\bar{h})}\neq y|H=h,Y=y)$ denotes the counterfactual probability of necessity. The $\mathrm{do}(\cdot)$ operator signifies an intervention on the variable $H$, distinguishing it from mere observation. A high PNS value provides a strong predictive relationship between $H$ and $Y$. However, collecting data on counterfactuals is often impossible in the real world. Fortunately, PNS becomes identifiable under two conditions: Exogeneity and Monotonicity \cite{pearl2009causality}.

Exogeneity holds when the intervention probability can be identified by the conditional probability: $P(Y_{\mathrm{do}(H=h)}=y) = P(Y=y|H=h)$ \cite{pearl2009causality}. This means the effect of setting $H$ to a particular value is the same as observing $H$ at that value. Monotonicity holds when the statement $(Y_{do(H=h)} \neq y) \wedge (Y_{do(H=\bar{h})} = y)$ is false or $(Y_{do(H=h)} = y) \wedge (Y_{do(H=\bar{h})} \neq y)$ is false\cite{pearl2009causality}. This can be represented as:
\begin{equation}
\label{eq:mono}
\begin{aligned}
P(Y_{do(H=h)} \neq y)P(Y_{do(H=\bar{h})} = y) = 0 
\,\, \text{or}\\
P(Y_{do(H=\bar{h})} = y)P(Y_{do(H=h)} \neq y) = 0
\end{aligned}
\end{equation}

Under these conditions, PNS can be calculated as\cite{pearl2009causality}:
\begin{equation}
\label{eq:pns}
\text{PNS}(Y, H) = P(Y=y|H=h) - P(Y=y|H=\bar{h})
\end{equation}

The proof of \cref{eq:pns} is provided in \cite{pearl2009causality, wang2021desiderata}. To apply \cref{eq:pns} to hidden features in our MIQA context, we make two assumptions. First, we assume Exogeneity of hidden features extracted by DL models, as these features are derived directly from the input and their relationship with the outcome is inherently conditional. Second, to ensure interpretable hidden features in PNS calculation, we assume that minor perturbations to the features preserve the same semantic meaning. Specifically, the features extracted by different DL models from the same input, or different features extracted by the same model from different inputs, exhibit semantic separability. This assumption is widely accepted to prevent unstable data issues, as similar feature values could otherwise correspond to entirely different semantics \cite{yang2023invariant}.

\begin{figure*}[t]
\centerline{\includegraphics[width=\textwidth]{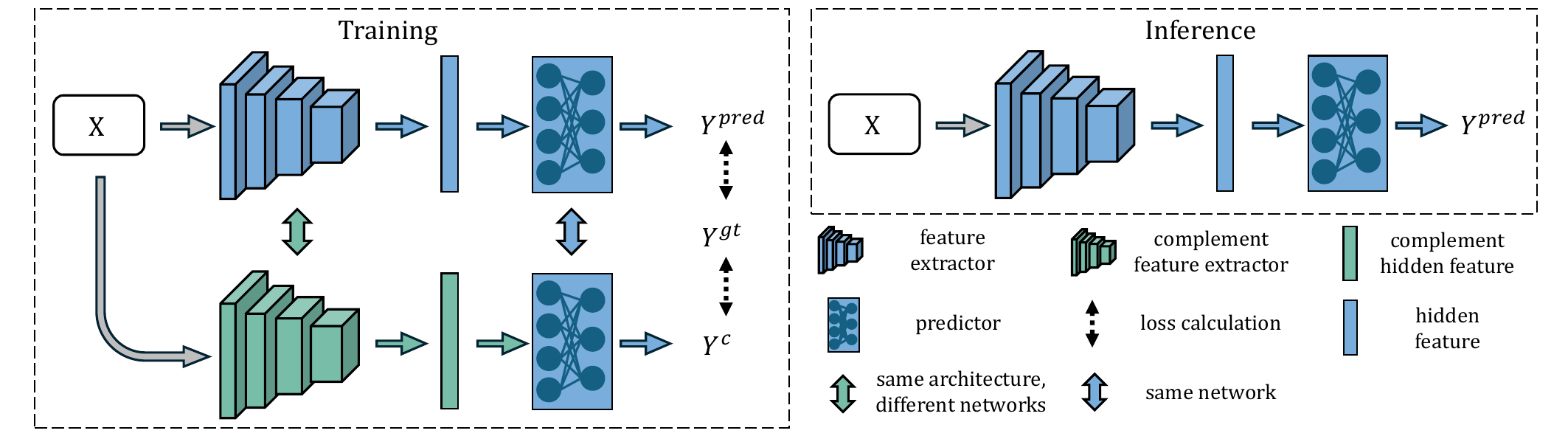}}
\caption{The architecture of our MIQA framework.}
\label{fig:framework}
\end{figure*}

% \begin{figure*}[t]
% \centerline{\includegraphics[width=16.5cm]{framework.pdf}}
% \caption{The architecture of our MIQA framework.}
% \label{fig:framework}
% \end{figure*}

\section{Method}
\label{sec:method}
We introduce MIQA-PNS, an MIQA framework to guide DL models to learn high-PNS features to distinguish between Good and Deficient images. As shown in \cref{fig:framework}, our framework consists of a feature extractor, a predictor, and a complement feature extractor. Inspired by \cite{yang2023invariant}, the objective function for MIQA-PNS is formulated as:
\begin{equation}
\label{eq:pns_loss}
\mathcal{L}^{task} :=  \mathcal{L}^{pred} + \mathcal{L}^{compl} + \mathcal{L}^{mono}
\end{equation}

\subsection{Prediction Loss $\mathcal{L}^{pred}$}
The prediction loss $\mathcal{L}^{pred}$ evaluates the performance of a DL model in the MIQA task. We treat one model as a combination of a feature extractor $\mathcal{E}$ and a predictor $\mathcal{F}$ (upper part of the training block in \cref{fig:framework}). Given an image $X$, $\mathcal{E}$ extracts hidden features $\mathcal{H} := \mathcal{E}(X)$, while $\mathcal{F}$ predicts the quality $Y^{pred} := \mathcal{F}(\mathcal{H})$. We then define $\mathcal{L}^{pred}$ as:
\begin{equation} \nonumber
\label{eq:pns_pred}
\mathcal{L}^{pred} :=  \mathcal{L}^{cls}(Y^{gt}, Y^{pred})
\end{equation}
where $\mathcal{L}^{cls}$ is a loss function that decreases as $Y^{pred}$ approaches the ground truth $Y^{gt}$. By optimizing this loss, we increase the probability of the prediction being close to $Y^{gt}$ given the hidden feature $\mathcal{H}$. This aligns with our goal of achieving a high $P(Y=y|H=h)$ in \cref{eq:pns}.

\subsection{Complement Loss $\mathcal{L}^{compl}$}
To evaluate PNS in \cref{eq:pns}, we need to obtain the complement $\bar{h}$ for feature value $h$ of $H$. We thus introduce a complement feature extractor $\mathcal{E}^{c}$ to extract $\mathcal{H}^{c}:= \mathcal{E}^{c}(X)$ for $\mathcal{H}$. $\mathcal{E}^{c}$ has a same structure as $\mathcal{E}$ but is a separate network. We then use $\mathcal{F}$ to predict $Y^{c}:=\mathcal{F}(\mathcal{H}^{c})$ that differ from $Y^{gt}$. The complement loss $\mathcal{L}^{compl}$ is defined as:

\begin{equation} \nonumber
\mathcal{L}^{compl} :=  \mathcal{I}(Y^{gt}) \cdot \mathcal{L}^{c}(Y^{gt}, Y^{c})
\end{equation}
where $\mathcal{L}^{c}$ is a loss function that increases when $Y^{c}$ is close to $Y^{gt}$. By optimizing it, we decrease the probability of the prediction being close to $Y^{gt}$ when the hidden feature is set to $\mathcal{H}^{c}$, mirroring the process of achieving a low $P(Y=y|H=\bar{h})$ in \cref{eq:pns}. The indicator function $\mathcal{I}(Y^{gt})$ equals 1 if $Y^{gt}$ represents Good, and 0 otherwise. This focus ensures that our framework learns high-PNS features specifically for good quality images.

\subsection{Monotonicity Constrain Loss $\mathcal{L}^{mono}$}
If Monotonicity holds, optimizing $\mathcal{L}^{pred} + \mathcal{L}^{compl}$ is the process of improving the PNS in \cref{eq:pns} for predicting good quality by using extracted features. We thus define the Monotonicity constraint loss as:
\begin{equation} \nonumber
\mathcal{L}^{mono} := \lambda \cdot \mathcal{I}(Y^{gt}) \cdot \mathcal{L}^{cls}(Y^{gt}, Y^{pred}) \cdot \mathcal{L}^{c}(Y^{gt}, Y^{c})
\end{equation}
where $\lambda$ controls the importance of the constraint and $\mathcal{I}(Y^{gt})$ focuses the framework on good quality images. By minimizing the product $\mathcal{L}^{cls}(Y^{gt}, Y^{pred}) \cdot \mathcal{L}^{c}(Y^{gt}, Y^{c})$, we encourage the extracted features to satisfy the condition $P(Y_{do(H=h)} \neq y)P(Y_{do(H=\bar{h})} = y)=0$ in \cref{eq:mono}. The multiplication of these probabilities decreases as $\mathcal{L}^{mono}$ decreases, thereby promoting the Monotonicity condition.

\subsection{Implementation}
We utilize four popular DL backbones as $\mathcal{E}$: VGG16 \cite{simonyan2014very}, ResNet50 \cite{he2016deep}, EfficientNet-B0 \cite{tan2019efficientnet}, and RegNet50 \cite{radosavovic2020designing}. We empirically set the $\lambda$ values for these models as 1.0, 0.6, 0.8, and 0.6, respectively. The $\mathcal{F}$ is implemented as a fully-connected network with hidden layers of sizes [256, 64].
We employ cross-entropy loss $\mathcal{L}^{CE}$ as our $\mathcal{L}^{cls}$.
To implement $\mathcal{L}^{c}$, we introduce a label transform function $T(\cdot)$, where $T(Y^{gt})$ produces a label different from $Y^{gt}$. We then define $\mathcal{L}^{c}(Y^{gt}, Y^{c}) := \mathcal{L}^{CE}(T(Y^{gt}), Y^{c})$. During the training process, we optimize $\mathcal{E}$, $\mathcal{E}^{c}$, and $\mathcal{F}$ simultaneously using the Adam optimizer with a learning rate of 0.0001. We implement an early stopping strategy with a patience of 15 epochs to prevent overfitting. After training, we discard $\mathcal{E}^{c}$, retaining only $\mathcal{E}$ and $\mathcal{F}$ for the inference stage (the right part of \cref{fig:framework}).

\section{Experiment and Results}
\label{sec:exp}

\subsection{Dataset}
Our experiments utilize the same AS-OCT image dataset from \cite{chen2023automated}, comprising 2,825 images categorized as Good (593), Limited (1,827), and Poor (405). We group Limited and Poor into Deficient category (\cref{fig:quality}) to focus on distinguishing Good quality images from the rest. We conduct two experiments: First, We demonstrate MIQA-PNS's ability to enhance DL models' MIQA performance using a 70\%/15\%/15\% train/validation/test split of the full dataset. Second, we test MIQA-PNS's OOD generalization capacity. Specifically, we design two scenarios: (1) training and validating (70\%:30\% split) on Good and Poor images while testing on Limited images (classified as Deficient), and (2) training and validating (70\%:30\% split) on Good and Limited images while testing on Poor images (classified as Deficient). These scenarios simulate conditions where certain types of Deficient images are unseen during training.

\subsection{Results and Discussion}
Table \ref{tab:exp1} presents the results of first experiment. For each model, we compare performance when optimizing only $\mathcal{L}^{pred}$ in \cref{eq:pns_loss} (original DL model) versus optimizing $\mathcal{L}^{task}$ (MIQA-PNS). The evaluation metrics are Precision, Recall and F1-score. The results demonstrate our framework can improve MIQA performance across all tested models.

Table \ref{tab:exp2} shows the results of second experiment, displaying the accuracy of predicting Deficient images in the testing set. Notably, accuracy is consistently high when predicting Deficient images in the Poor-only testing set, regardless of whether MIQA-PNS is used. We attribute this to the model learning that eyelashes are a predictive feature for Limited images, which is also present in most Poor images. 

In the Limited-only testing dataset, accuracy is consistently lower across all models due to the varying nature of image categories. While Poor images typically contain easily distinguishable eyelashes, Limited images present challenging, often sharing more similarities with Good images \cite{chen2023automated}. Some Limited images feature eyelashes (e.g., \cref{fig:quality}. c), allowing models trained on Good and Poor examples to classify them as Deficient. However, the absence of eyelashes in other Limited images (e.g., \cref{fig:quality}. d) makes differentiation from Good images more difficult. This is where MIQA-PNS provides value: by encouraging models to learn more necessary and sufficient predictive information for Good, it enhances model's performance in OOD scenarios. This suggests MIQA-PNS can help capture more robust features for image quality beyond simply relying on the presence of eyelashes.

While our results using AS-OCT images are promising, our study has a limited scope. Future research should explore the framework's applicability to a broader range of imaging modalities and quality assessment criteria to fully evaluate its generalizability. Despite this limitation, we believe that encouraging DL models to learn more necessary and sufficient predictive information has great potential. It not only provides novel insights for MIQA but also holds promise for other classification task in medical imaging.

\begin{table}
\caption{The performance for identifying Good image of DL models without and with using MIQA-PNS framework}
\label{tab:exp1}
\centering
\begin{tabular}{llll}
\hline
Model & Precision & Recell & F1 \\
\hline
VGG &  0.838 & 0.876 & 0.857  \\
VGG (MIQA-PNS) & \textbf{0.868} & \textbf{0.887} & \textbf{0.877}  \\
\hline
ResNet & 0.855 & 0.865 & 0.860  \\
ResNet (MIQA-PNS) & \textbf{0.867} & \textbf{0.882} & \textbf{0.875}  \\
\hline
EfficientNet & 0.868 & 0.890 &  0.879 \\
EfficientNet (MIQA-PNS) & \textbf{0.879} & \textbf{0.898} &  \textbf{0.888} \\
\hline
RegNet & 0.855 & 0.848 &  0.852 \\
RegNet (MIQA-PNS) & \textbf{0.856} & \textbf{0.899} & \textbf{0.877}  \\
\hline
\end{tabular}
\end{table}

\begin{table}
\caption{The accuracy for predicting Deficient in testing dataset that contains only Limited or Poor images}
\label{tab:exp2}
\centering
\begin{tabular}{lll}
\hline
Model & Only Limited & Only Poor\\
\hline
VGG & 0.166 &  1.0 \\
VGG (MIQA-PNS) & \textbf{0.195} & 1.0  \\
\hline
ResNet & 0.186 & 1.0  \\
ResNet (MIQA-PNS) & \textbf{0.203} & 1.0  \\
\hline
EfficientNet & 0.193 & 0.987  \\
EfficientNet (MIQA-PNS) & \textbf{0.216} &  \textbf{0.992} \\
\hline
RegNet & 0.159 & 1.0  \\
RegNet (MIQA-PNS) & \textbf{0.189} & 1.0  \\
\hline
\end{tabular}
\end{table}

\section{Conclusion}
\label{sec:conclusion}
We introduce a framework that encourages DL models to learn more reliable predictive information for MIQA tasks. Experiments demonstrate that this approach can enhance both the MIQA performance and robustness of those models.

% \bibliographystyle{IEEEtran}
% \bibliography{refs}

\end{document}